\documentclass[conference]{IEEEtran}
\IEEEoverridecommandlockouts
\usepackage{cite}
\usepackage{amsmath,amssymb,amsfonts}
\usepackage{algorithmic}
\usepackage{graphicx}
\usepackage{textcomp}
\usepackage{xcolor}

\usepackage{multirow}
\usepackage{algorithm}
\usepackage{url} 

\usepackage{booktabs}

\usepackage{tikz}

\newcommand{\first}[1]{\bf{#1}}
\newcommand{\second}[1]{\underline{#1}}
\newcommand{\tikzcmark}{%
\tikz[scale=0.23] {
    \draw[line width=1,line cap=round] (0.25,0) to [bend left=10] (1,1);
    \draw[line width=1,line cap=round] (0,0.35) to [bend right=1] (0.23,0);
}}

\newcommand{\tikzxmark}{%
\tikz[scale=0.23] {
    \draw[line width=1,line cap=round] (0,0) to [bend left=6] (1,1);
    \draw[line width=1,line cap=round] (0.2,0.95) to [bend right=3] (0.8,0.05);
}}

\def\BibTeX{{\rm B\kern-.05em{\sc i\kern-.025em b}\kern-.08em
    T\kern-.1667em\lower.7ex\hbox{E}\kern-.125emX}}
\begin{document}

\title{Harmonized Tabular-Image Fusion via Gradient-Aligned Alternating Learning}

\author{
  Longfei Huang$^1$, Yang Yang$^{1*}$\thanks{*Corresponding author.}\\
  $^1$Nanjing University of Science and Technology\\
  \{hlf, yyang\}@njust.edu.cn \\
}

\maketitle

\begin{abstract}
Multimodal tabular-image fusion is an emerging task that has received increasing attention in various domains. However, existing methods may be hindered by gradient conflicts between modalities, misleading the optimization of the unimodal learner. In this paper, we propose a novel Gradient-Aligned Alternating Learning (GAAL) paradigm to address this issue by aligning modality gradients. Specifically, GAAL adopts an alternating unimodal learning and shared classifier to decouple the multimodal gradient and facilitate interaction. Furthermore, we design uncertainty-based cross-modal gradient surgery to selectively align cross-modal gradients, thereby steering the shared parameters to benefit all modalities. As a result, GAAL can provide effective unimodal assistance and help boost the overall fusion performance. Empirical experiments on widely used datasets reveal the superiority of our method through comparison with various state-of-the-art (SoTA) tabular-image fusion baselines and test-time tabular missing baselines. The source code is available at \url{https://github.com/njustkmg/ICME26-GAAL}.
\end{abstract}


\section{Introduction}
\begin{figure}[t] 
\centering
\includegraphics[scale=0.3]{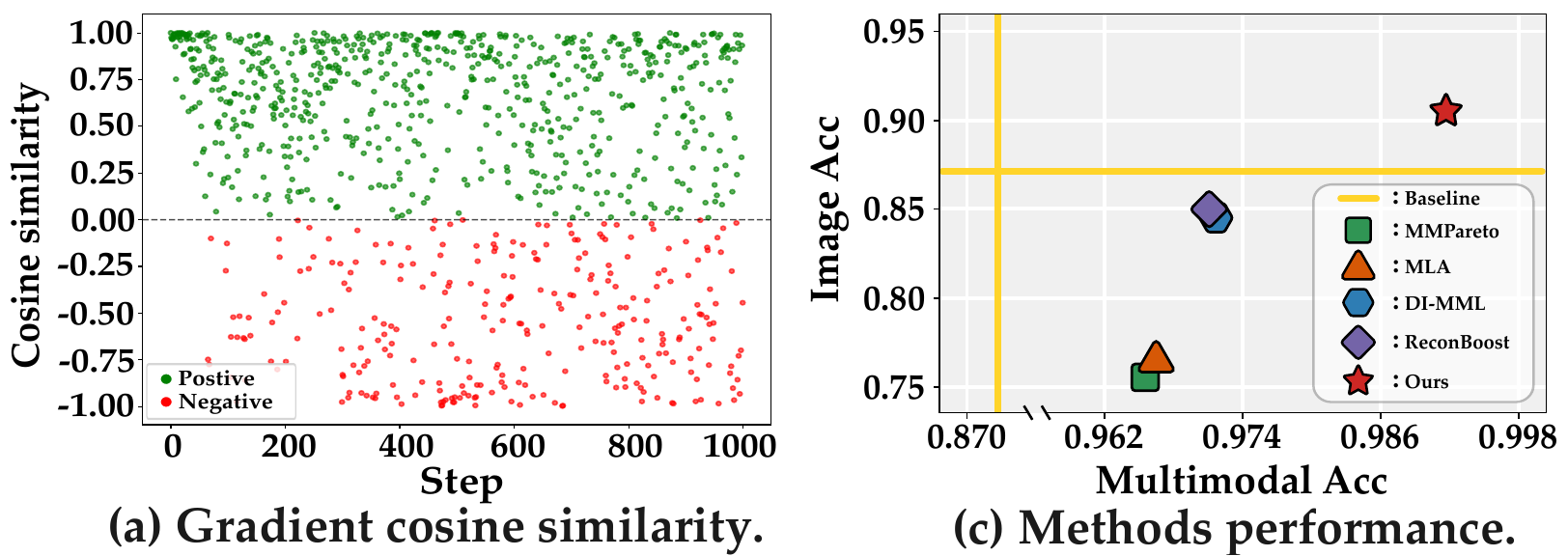}
\caption{We visualize the gradient conflicts and evaluate the performance of existing multimodal solutions on DVM dataset. (a) Multimodal and image gradients often show negative cosine similarity in naive joint learning, indicating severe gradient conflicts. (b) Existing multimodal solutions underperform in tabular-image fusion, failing to fully exploit unimodal image potential.}
\label{fig:imbalance}
\end{figure}

In recent years, tabular data is increasingly accessible in multimodal datasets, and its integration is crucial in various applications \cite{Recommender:conf/kdd/BaltescuCPZLR22,Healthcare:conf/kdd/ZhangCMZWWZ22,Healthcare:acosta2022multimodal}. An emerging example is tabular-image fusion that involves integrating structured tables and images to provide a holistic understanding of subjects. Despite numerous achievements, existing tabular-image fusion approaches \cite{DAFT:journals/neuroimage/WolfPWIa22,TIP:conf/eccv/DuZWBOQ24} often follow multimodal joint learning, where gradient conflicts frequently arise. These modality gradient conflicts are primarily caused by the unified optimization objective in joint learning \cite{KMG:conf/icmcs/PanY25,AUG:conf/nips/Jiang2025aug}. This potentially misleads unimodal learning and results in suboptimal final performance. As shown in Figure \ref{fig:imbalance}(a), we visualize the cosine similarity between image gradients and multimodal gradients on DVM \cite{DVM:conf/bigdataconf/HuangCLYO22} dataset. Negative cosine similarity indicates the presence of conflicts.

Fortunately, recent multimodal studies \cite{OGR-GB:conf/cvpr/WangTF20,KMG:journals/pami/YangPJXT25,MMPareto:conf/icml/WeiH24} have recognized this issue and attempted to provide solutions. An early representative work, OGM \cite{OGM:conf/cvpr/PengWD0H22}, balances strong and weak modalities by adjusting gradient magnitudes. Subsequent work, MMPareto \cite{MMPareto:conf/icml/WeiH24}, introduces Pareto methods from multi-task learning to regulate gradient directions. Beyond the above joint learning paradigm, other works \cite{MLA:conf/cvpr/ZhangYBY24,ReconBoost:conf/icml/CongHua24,DI-MML:conf/mm/FanXWLG24} adopt an alternating learning paradigm to decouple combined gradient into unimodal gradients, achieving higher performance. To facilitate interaction, they adopt a shared-head strategy.

However, the aforementioned methods mainly focus on alleviating gradient conflicts in unimodal encoders, at the expense of the facilitative role of gradient signals in cross-modal interaction layer. This issue becomes pronounced when directly transferred to more challenging tabular-image tasks. As shown in Figure \ref{fig:imbalance}(b), these methods suffer from significant performance degradation in the tabular-image fusion, with image performance even falling below that of unimodal learning. This can be attributed to two main issues. First, gradient modulation is inherently more challenging in joint learning, making it difficult to unleash unimodal potential effectively. Second, while alternating learning can liberate unimodal encoders, shared head suffers from gradient conflicts. MLA \cite{MLA:conf/cvpr/ZhangYBY24} uses gradient orthogonalization to encourage modal independence, neglecting overlaps and synergy among modal objectives.

In this work, to overcome the above challenges, we propose Gradient-Aligned Alternating Learning (GAAL), a novel tabular-image learning paradigm that effectively addresses gradient conflict and strengthens the tabular-image fusion performance. Specifically, GAAL alternates modality-specific optimization to decouple multimodal gradients while capturing cross-modal interaction via a shared classifier, thereby performing relatively independent unimodal training. Then, we design uncertainty-based cross-modal gradient surgery that modulates modal gradients via quadratic programming (QP) \cite{QP:frank1956algorithm,GEM:conf/nips/Lopez-PazR17} to address gradient conflicts and synergy at shared classifier. This surgery computes gradients from high-entropy samples of the previous modality and projects the current modality gradient onto the previous gradient direction while minimizing their Euclidean distance. To this end, GAAL addresses gradient conflicts and facilitates interactions in tabular-image fusion. As shown in Figure \ref{fig:imbalance}(b), our proposed method GAAL improves both multimodal and unimodal performance. Our main contributions are outlined as follows:
\begin{itemize} 
\item We propose GAAL, a novel alternating learning algorithm that allows the model to explore unimodal information effectively, thereby improving tabular-image fusion. 
\item We design an uncertainty-based cross-modal gradient surgery that guides modal optimization directions to address gradient conflicts and facilitate synergy at the interaction layer.
\item Experimental results demonstrate that GAAL effectively addresses gradient conflict to improve tabular-image fusion performance and achieves SoTA performance.
\end{itemize}
\begin{figure*}
\centering
\includegraphics[width=0.7\textwidth]{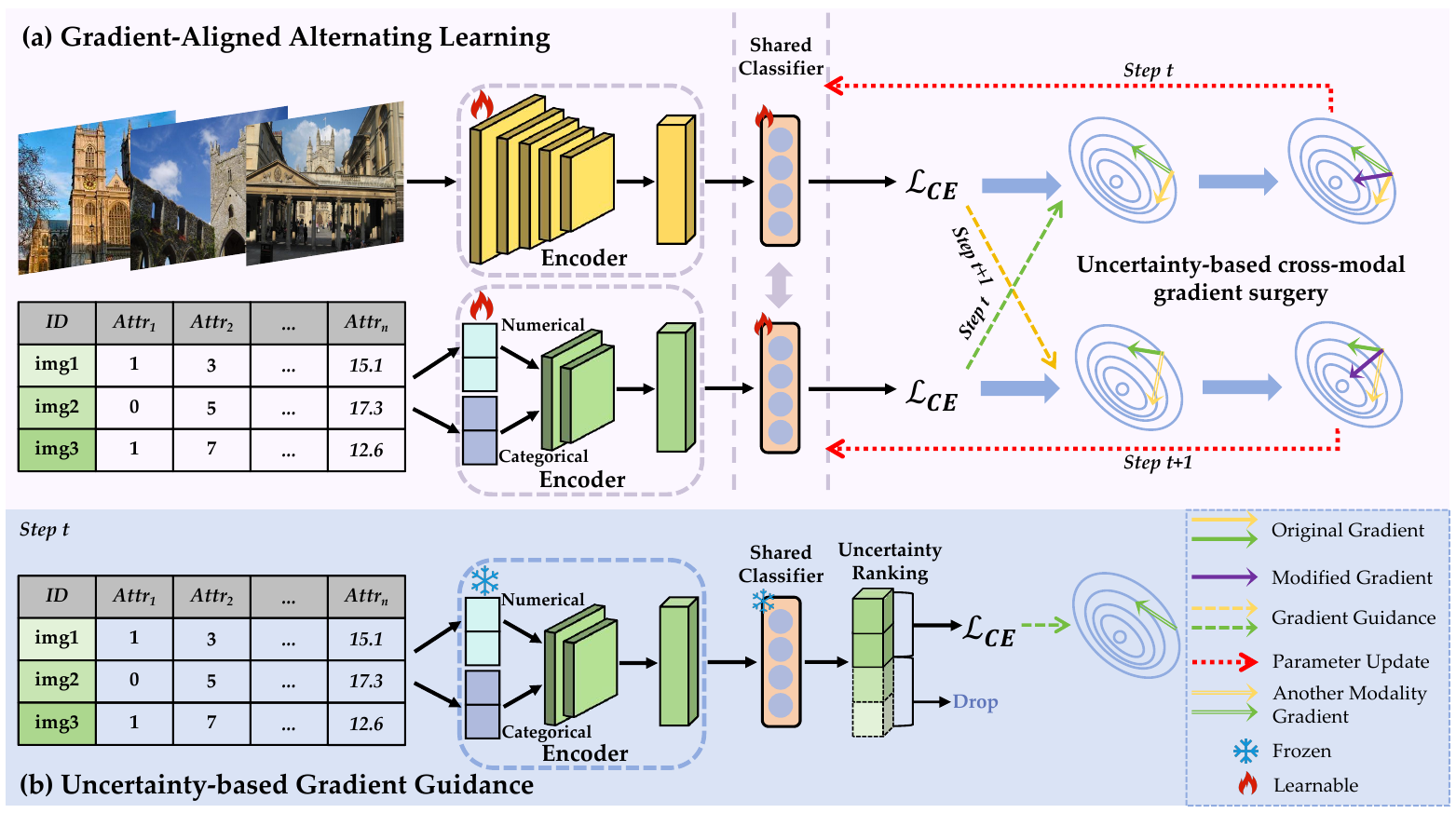}
\caption{The framework of the GAAL method. (a) Gradient-Aligned Alternating Learning that only one modality's learner is updated at each step. Uncertainty-based cross-modal gradient surgery utilizes gradients from cross-modal hard samples to guide the optimization of the shared classifier for the current modality. (b) Uncertainty-based Gradient Guidance that samples hard examples from another modality to provide cross-modal gradient guidance.}
\label{fig:framework}
\end{figure*}

\section{Related Work}
\subsection{Tabular-image Fusion}
Tabular data, known for its structured and interpretable nature, is the focus of traditional machine learning and deep learning \cite{Tabular:conf/nips/GorishniyRKB21,Tabular:conf/nips/MargeloiuJSJ24}. Unlike free-form audio or textual data, structured tabular data mixes dense numerical and sparse categorical features with differing value ranges and semantics, and lacks clearly defined interrelationships \cite{Tabular:journals/tnn/BorisovLSHPK24}. Existing tabular-image fusion approaches focus on integrating cross-modal information by multimodal learning strategies. TIP \cite{TIP:conf/eccv/DuZWBOQ24} proposes a transformer-based table encoder and enhances the robustness of tabular-image models through contrastive learning and cross-attention fusion. Moreover, since obtaining real-world tabular data is often costly and challenging, other studies have explored settings where tabular data are unavailable at inference time. These methods transfer expensive tabular expertise to images during training to enhance the performance of image models at inference. Since our method can facilitate cross-modal interaction, it also maintains good performance even when test-time tabular data are unavailable.

\subsection{Multimodal Gradient Conflict}
Gradient conflict, defined as negative cosine similarity between gradients, has been extensively studied in multiple domains \cite{KMG:conf/ijcai/JiangCY25,Nash:conf/iccv/Wu_2025_ICCV}. Recent studies \cite{OGR-GB:conf/cvpr/WangTF20,MMPareto:conf/icml/WeiH24} have found that gradient conflicts also exist in multimodal scenarios. Such multimodal gradient conflicts can mislead unimodal optimization, ultimately resulting in suboptimal multimodal performance. To address this issue, some early works \cite{OGR-GB:conf/cvpr/WangTF20,OGM:conf/cvpr/PengWD0H22} adjust the magnitude of modality-specific gradients based on unimodal performance to compensate for weak modality. Subsequent work MMPareto \cite{MMPareto:conf/icml/WeiH24} introduces Pareto approach in multi-task learning to identify integrated gradient directions beneficial for all modalities. Moreover, since multimodal gradient conflicts mainly arise from the unified optimization objective in joint learning, recent methods adopt an alternating learning paradigm to decouple optimization objective. Although this framework can directly address gradient conflicts in the encoder, shared interaction layers across modalities still suffer from gradient conflicts. Notably, MLA also considers interaction layer, but it isolates modal gradients through orthogonalization, overlooking potential overlap and synergy among modal optimization objectives. Therefore, our method aims to address both gradient conflicts and cross-modal interaction.

\section{Methodology}
In this section, we introduce the proposed GAAL, a tabular-image alternating learning framework that addresses gradient conflict to improve classification. To facilitate tabular-image interaction, we design an uncertainty-guided cross-modal gradient surgery, which utilizes uncertainty-based cross-modal gradient guidance to assist in optimizing unimodal samples. The architecture of GAAL is shown in Figure \ref{fig:framework}. 

\subsection{Preliminary}
Let $\mathcal{X}= {\left \{ {x}_{i}^{I}, {x}_{i}^{T}, {y}_{i}\right \}}_{i=1}^{N}$ be a training set, where ${y}_{i}$ is the label for the ${i}$-th instance and $N$ is the number of training data. ${x}_{i}^{I}\in {\mathbb{R}}^{H\times W\times 3}$ represents the image, and ${x}_{i}^{T}\in {\mathbb{R}}^{D}$ represents the tabular description, where $D$ is the number of tabular features. Each tabular input contains two kinds of attributes, i.e., the categorical features (such as the ``Gender'') and the continuous features (such as the ``Age''). Assume there are
$Y$ classes in total, and ${y}_{i}\in \left [ Y\right ]=\left \{ 1,...,Y\right \}$. For the sake of simplicity, we use superscript $m$ to indicate the module corresponding to a specific modality in this section, where $m \in \left \{ I, T\right \}$. Notation is summarized in supplemental material.

\subsection{Gradient-Aligned Alternating Learning}
With the rapid growth of deep learning, representative tabular-image fusion approaches \cite{MMCL:conf/cvpr/HagerMR23,TIP:conf/eccv/DuZWBOQ24,CHARMS:conf/icml/JiangYW00Z24} have adopted deep neural network (DNN) for multimodal learning. Following these methods, we utilize DNN to construct our tabular and image models. Specifically, We use $\phi^m$ as encoders to extract features ${u}^{m}= {\phi }^{m}({\theta }^{m}, {x}^{m})$, where $\theta $ the encoder parameters. We define the predictor $\psi$ as a mapping from the latent feature space to the label space. For given tabular-image pairs $x=\left [x^I, x^T \right ]$, the tabular-image model can be written as $f(x)=\psi (\left [{u}^{I}:{u}^{T} \right ])$. Therefore, the objective function can be written as:
\begin{align}
{\mathcal{L}}(x, y)=-\frac{1}{N}\displaystyle\sum_{i=1}^{N}{y}_{i}^{\top}\log{({f(x)}_{i})}.
\label{obj:CE}
\end{align}

For a given input data instance, the alternating paradigm selects and updates a specific modality learner $f^m(x^m)=\psi ({\varTheta }^m, {\phi }^{m}(x^m)$ at each time, where $\psi $ is a shared classifier and $\varTheta $ denotes the parameters of the classifier. The objective function ${\mathcal{L}}^m(x^m, y)$ of each modality is independent. By utilizing alternating learning, we can successfully address gradient conflicts in the tabular and image encoders. Since interaction layer is shared, gradient conflicts may arise in shared head during alternating learning, overshadowing the previous modality’s optimization. Hence, to address this issue, we propose a cross-modal gradient surgery that adjusts the current gradients based on cross-modal gradients.

Concretely, assuming the gradient of the current modality at step $t$ is denoted as $g$, and the gradient from the previous modality is $g_p$. We define $\alpha $ as the angle between $g$ and $g_p$. We consider $g$ to be detrimental to the previous modality when $\cos \alpha < 0$. The cosine similarity $\mathcal{S}(\cdot)$ between two gradients $g$ and $g_p$ is  
\begin{align}
\mathcal{S}(g,g_p)=\frac{g^{\top}g_p }{\left \| g\right \|_2  \left \| g_p\right \|_2},
\label{obj:similarity}
\end{align}
where $\left \|\cdot  \right \|_2$ denote the $L_2$ norm of vectors.

We aim to project $g$ with the constraint to obtain the modified gradient $\widetilde{g}$ that is aligned with $g_p$:
\begin{align}
{g}_{p}^{\top}{\widetilde{g}}\geqslant \epsilon,
\label{obj:proj}
\end{align}
where $\epsilon$ is a small constant that governs the alignment intensity. Inspired by \cite{GEM:conf/nips/Lopez-PazR17,AGEM:conf/iclr/ChaudhryRRE19}, we can formulate (\ref{obj:proj}) as the following optimization problem:
\begin{align}
\begin{matrix}
\min_{\widetilde{g}} &{\frac{1}{2}{\left \| \widetilde{g}-g\right \|}^{2}_2}\\ 
s.t.&{g}_{p}^{\top}{\widetilde{g}}\geqslant \epsilon.
\end{matrix}
\label{obj:opt}
\end{align}

Since the objective function is a convex quadratic function and the constraint is a linear inequality, the problem reduces to a convex quadratic programming problem with a single constraint. Geometrically, $\widetilde{g}$ is the projection of $g$ onto $g_p$, and $\widetilde{g}$ is the closest to $g$ in Euclidean distance.

We introduce a dual variable $v$ to represent the gradient weight. By constructing the Lagrangian function and applying KKT conditions, we derive both the closed-form optimal solution as follows: 
\begin{equation}
\begin{aligned}
&\widetilde{g}=g + v{g}_{p}, \\
&v=\max (0, \frac{\epsilon -{g}_{p}^{\top}{g}}{{\left \| g_p\right \|}^{2}_2}).
\label{obj:solution}
\end{aligned}
\end{equation}
Unlike forcibly orthogonalizing all modal gradients, our method targets only conflicting gradients without enforcing orthogonality, which promotes multimodal synergy. More details are provided in supplemental material.

During alternating iterations, we use cross-modal gradients to guide current modality, which addresses gradient conflicts in the interaction layer and improves cross-modal interaction.

\subsection{Uncertainty-based Gradient Guidance}
Through the above gradient surgery, we successfully address gradient conflicts in the interaction layer and improve cross-modal interaction. Subsequently, we aim to select cross-modal gradients to provide more effective guidance in gradient surgery. Since gradients corresponding to hard samples often exhibit larger magnitudes, encapsulate more informative learning signals, and play a more critical role in parameter updates. Therefore, we design Uncertainty-based Gradient Guidance, which selects the gradient direction of high-uncertainty cross-modal samples as the projection direction. This gradient selection not only enables cross-modal gradients to guide current modality  but also utilizes current modality to assist in learning cross-modal samples. Learning hard samples effectively enhances the model's attention to edge cases.

Specifically, for a sample ${x}_{i}^{m}$ in a batch, let ${p}_{i}^{m}$ denote the model’s predicted class probability distribution. We evaluate the uncertainty of each sample by computing the entropy of its predicted probability distribution ${p}_{i}^{m}$ within the batch:
\begin{align}
\mathcal{H}(p_i^m)=-\displaystyle\sum_{c=1}^{Y}{p}_{i,c}^{m}\log{{p}_{i,c}^{m}},
\label{obj:ent}
\end{align}
where $\mathcal{H}(p_i^m)$ is the probability entropy for sample $i$.

Finally, we select the top of ${\lambda}^m$ samples with the highest uncertainty from modality $m$ as hard samples to compute the gradient $g_p$.
\begin{align}
\mathcal{I}_p^m = \operatorname{Top}_{\lambda^m}\left( \mathcal{H}(p_i^m) \mid i \in \mathcal{B}^m \right),
\label{obj:select}
\end{align}
\begin{align}
\quad
g_p = \frac{1}{|\mathcal{I}_p^m|} \sum_{i \in \mathcal{I}_p^m} \nabla_{\theta} \mathcal{L}(x_i^m),
\label{obj:grad}
\end{align}
where $\mathcal{B}^m$ denotes the batch of modality $m$, $\mathcal{I}_p^m$ is the set of indices of the selected top-${\lambda}^m$ high-entropy samples, $\mathcal{L}(x_i^m)$ is the loss for sample $i$, and $\nabla_{\theta}$ represents the gradient with respect to classifier parameters. By selecting cross-modal gradients from hard samples, the learning signal can be effectively focused, reducing dilution from easy samples and helping to avoid local optima.

Our algorithm is summarized in Algorithm \ref{algo:ours}. To sum up, our method GAAL employs alternating learning with shared-head interaction. Subsequently, we aim to address gradient conflicts and enhance synergy in interaction layer by projecting conflicting unimodal gradients. GAAL first selects hard samples from the previous modality to compute cross-modal gradients. Before updating the interaction layer for the current modality, conflicting gradients are projected onto the cross-modal gradients. This promotes hard sample learning and cross-modal interaction while addressing gradient conflicts.

\begin{algorithm}[t] 
\caption{GAAL Algorithm.}\label{algo:ours}
\begin{algorithmic}[1]
\STATE {\bfseries Input:} Training set $\mathcal{X}$.
\STATE {\bfseries Output:} The learned DNN models for all modalities.
\textbf{INIT} Initialize iteration $t=1$. Initialize encoder parameters $\theta^I$, $\theta^T$ and shared classifier parameters $\varTheta $.
\REPEAT
\IF{$mod(t,2)=1$}
\STATE Sample $\forall \left \{x_i^I,x_i^T, y_i\right \}\in \mathcal{X}$,
\ENDIF
\STATE Pick up a specific modality $m\in \left [I, T \right ]$ in order;
\STATE Calculate predictions $p_{i}^I$ and $p_{i}^T$ in (\ref{obj:pred});
\STATE Calculate modality $m$ loss $\mathcal{L}^m$ in (\ref{obj:CE});
\STATE Update modality-specific encoder parameters $\theta_t^m$;
\STATE Calculate modality $m$ classifier gradient $g$;
\STATE Calculate sample entropy $\mathcal{H}^{(I,T)/ m}$ in (\ref{obj:ent});
\STATE Select the top of $\lambda ^{(I,T)/ m}$ high-entropy samples;
\STATE Calculate modality $(I,T)/ m$ classifier gradient ${g}_{p}$;
\STATE Calculate modified gradient $\widetilde{g}$ in (\ref{obj:solution});
\STATE Update shared parameters $\varTheta $ using $\widetilde{g}$;
\STATE Update $t=t+1$;
\UNTIL{Converge or reach maximum iterations.}
\end{algorithmic}
\end{algorithm}

\subsection{Model Inference}
During the inference phase, given an tabular-image pair $x_i=\left [ {x}_{i}^{I}, {x}_{i}^{T}\right ]$, we compute the corresponding unimodal logits ${f}_{i}^{m}({x}_{i}^{m})$. 
The unimodal prediction $p_i^m$ is:
\begin{align}
{p}_{i}^{m}={Softmax({f}_{i}^{m}({x}_{i}^{m}))},
\label{obj:unipred}
\end{align}
and the final multimodal prediction $p_i$ is then derived through a weighted average of these unimodal logits:
\begin{align}
{p}_{i}={Softmax(\frac{1}{2}({f}_{i}^{I}({x}_{i}^{I})+{f}_{i}^{T}({x}_{i}^{T})))}.
\label{obj:pred}
\end{align}

With harmonized tabular-image fusion, inference merely requires simple averaging fusion at the decision level to achieve superior prediction results.

\begin{table*}[ht]
\centering
% \vspace{-3mm}
\caption{The accuracy results on DVM, SUNAttribute, and CelebA datasets. The best performance is bolded, and the second-best is underlined. The image performance of test-time tabular missing methods represents their multimodal performance. The ``*'' indicates the version that follows the same pretrained weight setting as test-time tabular missing methods.}
\label{tab:main-exp}
\begin{tabular}{lccc ccc ccc}
\toprule
\multirow{2}{*}{\textbf{Method}} & \multicolumn{3}{c}{\textbf{DVM}} & \multicolumn{3}{c}{\textbf{SUNAttribute}} & \multicolumn{3}{c}{\textbf{CelebA}} \\
\cmidrule(lr){2-4} \cmidrule(lr){5-7} \cmidrule(lr){8-10}
 & Multi & Image & Tabular & Multi & Image & Tabular & Multi & Image & Tabular \\
\midrule
\multicolumn{10}{l}{\textit{Unimodal Learning}} \\
\midrule
Resnet50   & -      & \second{0.8743} & -      & -      & \second{0.8361} & -      & -      & \second{0.8146} & -      \\
MLP        & -      & -      & 0.8742 & -      & -      & 0.8082 & -      & -      & 0.7775 \\
\midrule
\multicolumn{10}{l}{\textit{Tabular-image Fusion}} \\
\midrule
CF         & 0.8793 & 0.0058 & 0.8792 & 0.8278 & 0.5544 & 0.8298 & 0.7740 & 0.5654 & 0.7723 \\
MF         & 0.8993 & -      & -      & 0.8333 & -      & -      & 0.7941 & -      & -      \\
DAFT       & 0.9460 & -      & -      & 0.8456 & -      & -      & 0.8202 & -      & -      \\
TIP        & 0.9545 & -      & -      & \second{0.8612} & -      & -      & \second{0.8224} & -      & -      \\
\midrule
\multicolumn{10}{l}{\textit{Multimodal Gradient Conflict}} \\
\midrule
OGM        & 0.8778 & 0.0076 & 0.8780 & 0.8305 & 0.6457 & \second{0.8312} & 0.7771 & 0.5328 & 0.7756 \\
MMPareto   & 0.9658 & 0.7549 & \second{0.8851} & 0.8475 & 0.7992 & 0.8243 & 0.8082 & 0.7941 & 0.7702 \\
MLA        & 0.9668 & 0.7664 & 0.8505 & 0.8417 & 0.8047 & 0.8194 & 0.8134 & 0.7915 & \second{0.7913} \\
DI-MML     & 0.9719 & 0.8499 & 0.8748 & 0.8475 & 0.8003 & 0.8212 & 0.8199 & 0.8130 & 0.7839 \\
ReconBoost & 0.9714 & 0.8499 & 0.8756 & 0.8498 & 0.8117 & 0.8224 & 0.8149 & 0.8133 & 0.7836 \\
LFM        & \second{0.9731} & 0.8543 & 0.8848 & 0.8487 & 0.8075 & 0.8222 & 0.8115 & 0.8110 & 0.7613 \\
\midrule
\textbf{GAAL}       & \first{0.9917} & \first{0.9057} & \first{0.9191} & \first{0.8668} & \first{0.8452} & \first{0.8368} & \first{0.8273} & \first{0.8222} & \first{0.7922} \\
\midrule
\multicolumn{10}{l}{\textit{Test-time Tabular Missing}} \\
\midrule
KD         & - & 0.8390 & -      & - & 0.8382 & -      & - & 0.8118 & -      \\
MFH        & -      & -      & -      & - & 0.8312 & -      & - & 0.7507 & -      \\
FMR        & - & 0.8427 & -      & - & 0.8347 & -      & - & 0.8003 & -      \\
MMCL       & - & 0.8203 & -      & - & 0.8431 & -      & - & 0.8041 & -      \\
CHARMS & - & \second{0.9175} & - & - & \second{0.8661} & - & - & \second{0.8220} & - \\
\midrule
\textbf{GAAL*} & - & \textbf{0.9358} & - & - & \textbf{0.8662} & - & - & \textbf{0.8313} & - \\
\bottomrule
\end{tabular}
\end{table*}

\section{Experiments}
\subsection{Experimental Settings}
\noindent\textbf{Dataset.} We conducted experiments on three datasets: Data Visual Marketing (DVM) \cite{DVM:conf/bigdataconf/HuangCLYO22}, SUNAttribute \cite{SUN:journals/ijcv/PattersonXSH14}, and CelebA \cite{CelebA:conf/iccv/LiuLWT15}. \textbf{DVM} contains 1,451,784 car images paired with tabular data. Following previous work \cite{MMCL:conf/cvpr/HagerMR23}, car models with less than 100 samples were removed, resulting in 286 target classes. \textbf{SUNAttribute} is constructed from 717 SUN dataset categories, each annotated with 20 scene attributes. \textbf{CelebA} is a facial attribute dataset containing 202,599 face images. More details are provided in supplemental material.

\noindent\textbf{Baselines and Evaluation Metric.} We selected various SoTA baselines for comparison, including tabular-image fusion methods: CF \cite{CF:spasov2019parameter}, MF \cite{MF:vale2021long}, DAFT \cite{DAFT:journals/neuroimage/WolfPWIa22}, TIP \cite{TIP:conf/eccv/DuZWBOQ24}; test-time tabular missing methods: KD \cite{KD:journals/corr/HintonVD15}, MFH \cite{MFH:conf/iclr/XueGRZ23}, FMR \cite{FMR:conf/aaai/YangZFJZ17}, MMCL \cite{MMCL:conf/cvpr/HagerMR23}, CHARMS \cite{CHARMS:conf/icml/JiangYW00Z24}; multimodal gradient conflict methods: OGM \cite{OGM:conf/cvpr/PengWD0H22}, MMPareto \cite{MMPareto:conf/icml/WeiH24}, MLA \cite{MLA:conf/cvpr/ZhangYBY24}, DI-MML \cite{DI-MML:conf/mm/FanXWLG24}, ReconBoost \cite{ReconBoost:conf/icml/CongHua24} and LFM \cite{LFM:conf/nips/0074WJ024}. Following the setting of CHARMS \cite{CHARMS:conf/icml/JiangYW00Z24}, we adopt accuracy as the evaluation metric. 

\noindent\textbf{Implementation Details.} Following MMCL \cite{MMCL:conf/cvpr/HagerMR23}, we employ a ResNet50 \cite{ResNet:conf/cvpr/HeZRS16} as image encoder and MLP as the tabular encoder. Additionally, we perform a grid search for hyperparameters and employ early stopping to select the best model. All experiments are conducted on an NVIDIA RTX A6000. More details are provided in supplemental material.

\subsection{Experimental Results}
To demonstrate the effectiveness of GAAL, we compare it with multiple popular methods on the three datasets shown in Table \ref{tab:main-exp}, with results reported as mean values. The ``-'' indicates that the corresponding method failed to produce results for the modality. In particular, the MFH \cite{MFH:conf/iclr/XueGRZ23} method fails to handle the complex multi-class classification tasks of the DVM dataset. In summary, from Table \ref{tab:main-exp}, we can observe that: 1) Compared with unimodal learning, tabular-image fusion methods, multimodal gradient conflict methods and test-time tabular missing methods, GAAL can achieve better performance in almost all cases. 2) GAAL can outperform existing SoTA baselines, achieving not only the best multimodal performance but also the highest unimodal accuracy. 3) The accuracy of test-time tabular missing setting demonstrates that our method achieves the best performance and remains robust. More results are provided in supplemental material.

\begin{figure}[t] 
\centering
\begin{minipage}{.45\linewidth}
\centering
\includegraphics[width=\linewidth]{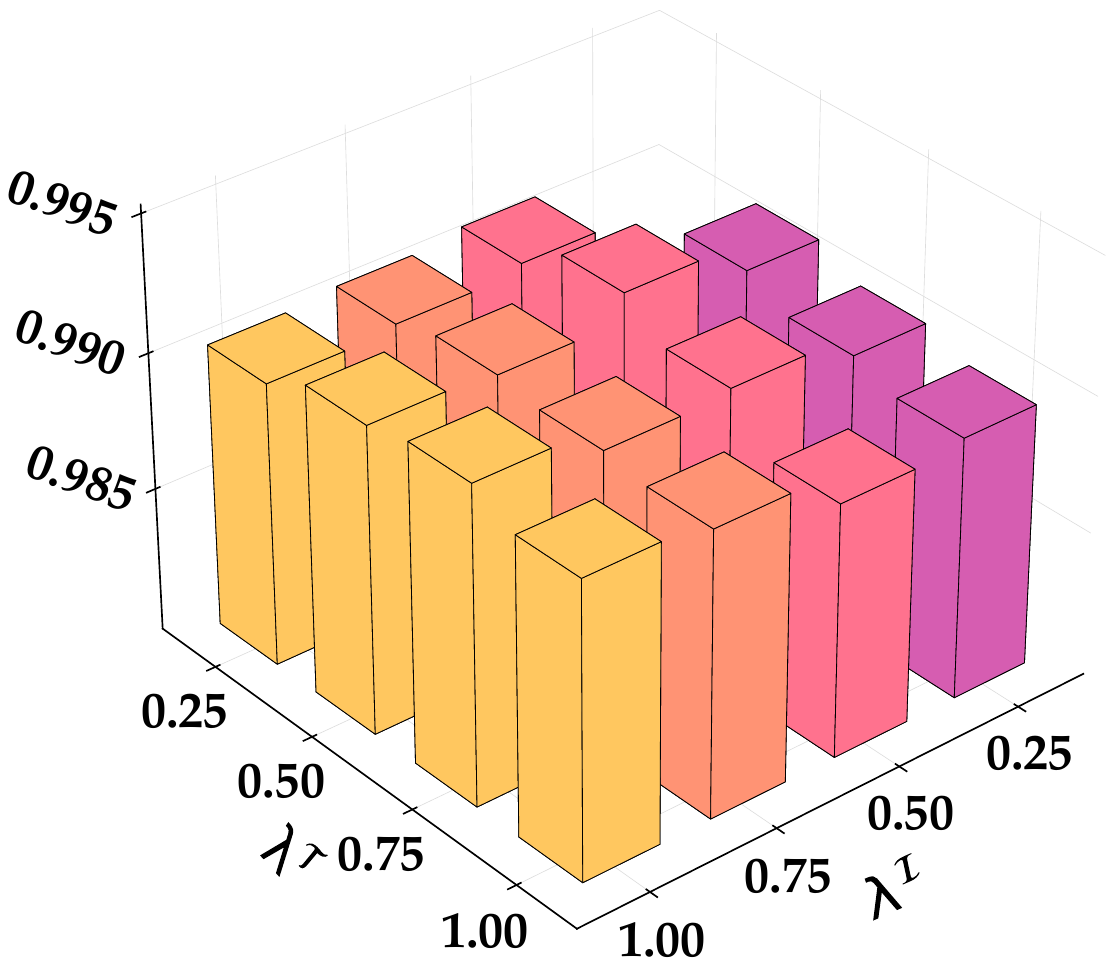}\\
{(a). DVM}
\end{minipage} 
\begin{minipage}{.45\linewidth}
\centering
\includegraphics[width=\linewidth]{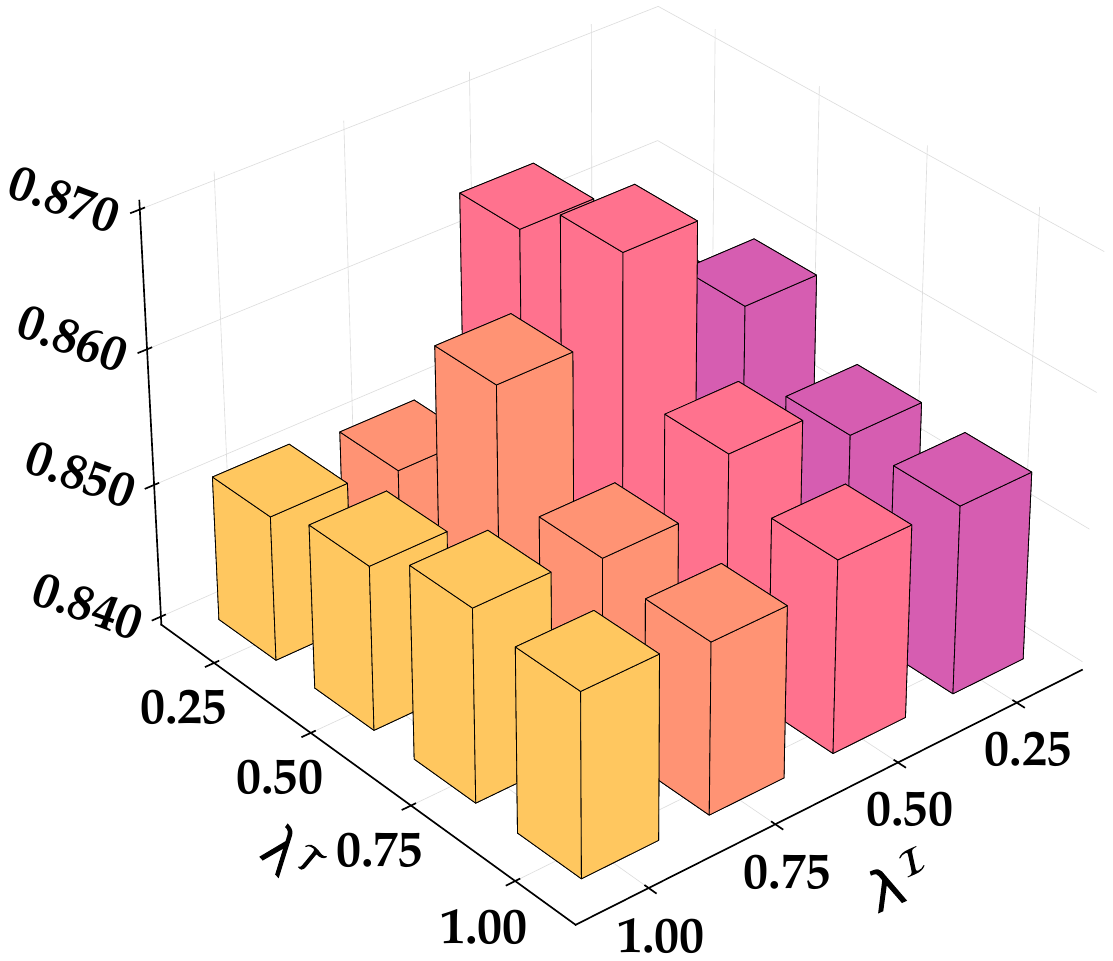}\\
{(b). SUNAttribute}
\end{minipage}
\caption{Sensitivity to hyper-parameter $\lambda ^I$ and $\lambda ^T$ on DVM and SUNAttribute datasets.}
\label{fig:sensitivity}
\end{figure}

\begin{table}[ht]
\centering
\caption{Results of ablation studies on DVM and SUNAttribute datasets.}
\label{tab:ablation}
\renewcommand{\arraystretch}{1.2}
\setlength{\tabcolsep}{1.5pt}
\begin{tabular}{cc|ccc|ccc}
\hline
\multirow{2}{*}{CGS} & \multirow{2}{*}{UGG} & \multicolumn{3}{c|}{\textbf{DVM}} & \multicolumn{3}{c}{\textbf{SUNAttribute}} \\
\cline{3-8}
                     &                      & Multi & Image & Tabular & Multi & Image & Tabular \\
\hline
 \tikzxmark & \tikzxmark & 0.9701 & 0.8552 & 0.8760 & 0.8375 & 0.7789 & 0.8103 \\
\tikzcmark & \tikzxmark & \second{0.9909} & \second{0.8931} & \first{0.9203} & \second{0.8536} & \second{0.8201} & \first{0.8459} \\
\tikzcmark & \tikzcmark & \first{0.9917} & \first{0.9057} & \second{0.9191} & \first{0.8668} & \first{0.8452} & \second{0.8368} \\
\bottomrule
\end{tabular}
\end{table}

\subsection{Sensitivity to Hyper-Parameters}
We study the impact of the threshold $\left \{\lambda ^I, \lambda ^T\right \}$ on GAAL performance, as shown in Figure \ref{fig:sensitivity}. An appropriate $\left \{\lambda ^I, \lambda ^T\right \}$ can effectively concentrate gradient signals, providing efficient cross-modal guidance for model optimization. Excessively large or small ${\lambda^I, \lambda^T}$ can have negative effects, either diluting gradient signals or ignoring data distribution.

\begin{figure}[t] 
\centering
\begin{minipage}{.45\linewidth}
\centering
\includegraphics[width=\linewidth]{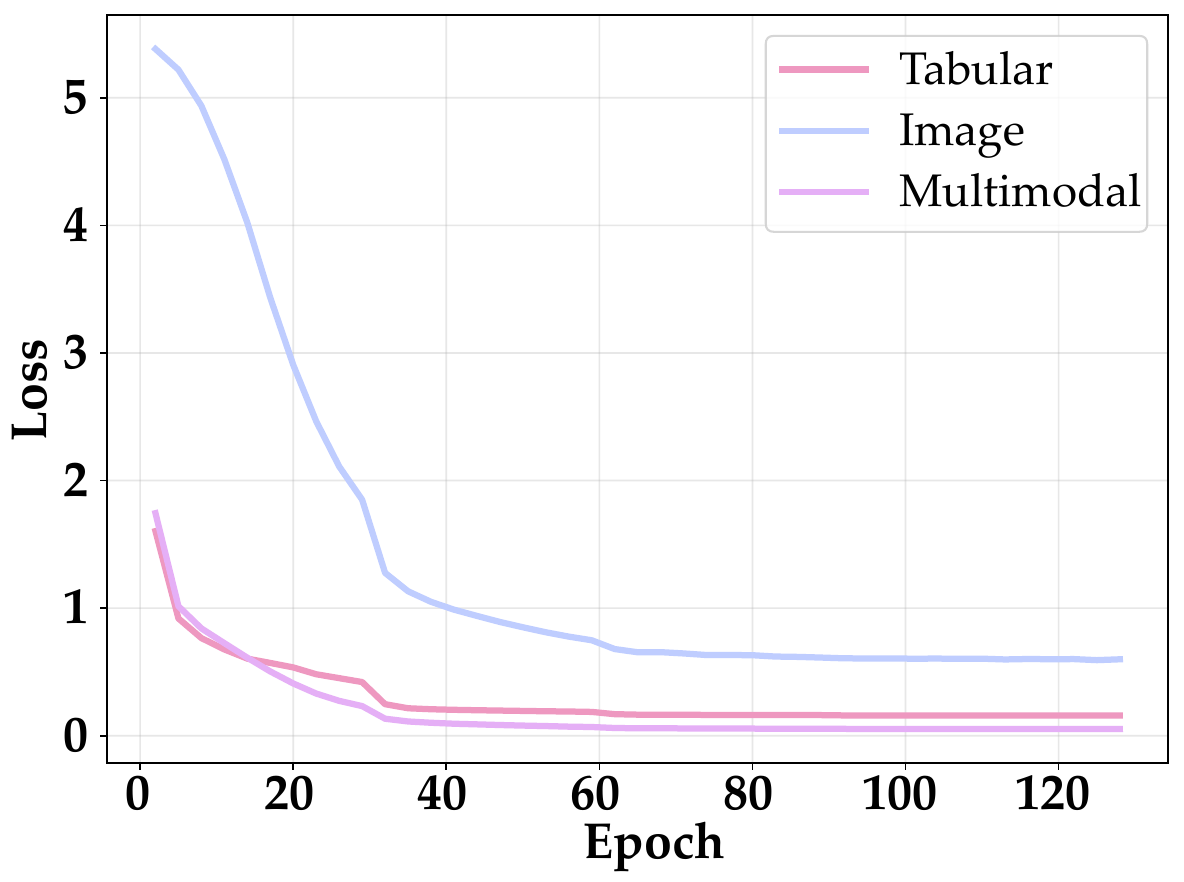}\\
{(a). DVM}
\end{minipage} 
\begin{minipage}{.45\linewidth}
\centering
\includegraphics[width=\linewidth]{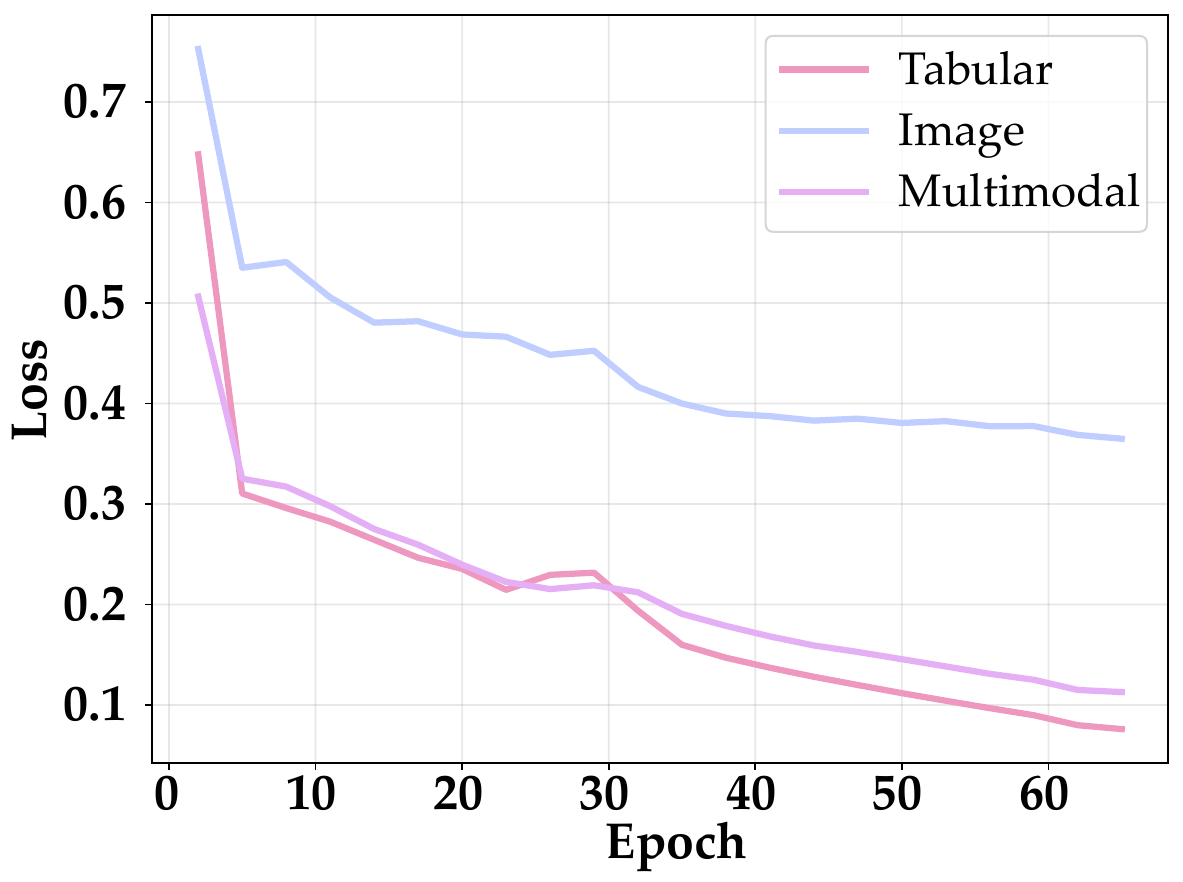}\\
{(b). SUNAttribute}
\end{minipage}
\caption{Convergence results of GAAL on DVM and SUNAttribute datasets.}
\label{fig:convergence}
\end{figure}

\subsection{Ablation Study}
We perform ablation studies on the DVM and SUNAttribute datasets to analyze the impact of alternating learning, cross-modal gradient surgery (CGS), and uncertainty-based gradient guidance (UGG), demonstrating the effectiveness of our method. Experimental results are reported in Table \ref{tab:ablation}. From Table \ref{tab:ablation}, we can find that: 1) Alternating learning, cross-modal gradient surgery and uncertainty-based gradient guidance can boost multimodal performance in terms of accuracy. 2) While the unimodal performance of the method using all objectives may not always reach the highest level, it achieves a more balanced classification performance across modalities.

\begin{figure}[t] 
\centering
\includegraphics[scale=0.3]{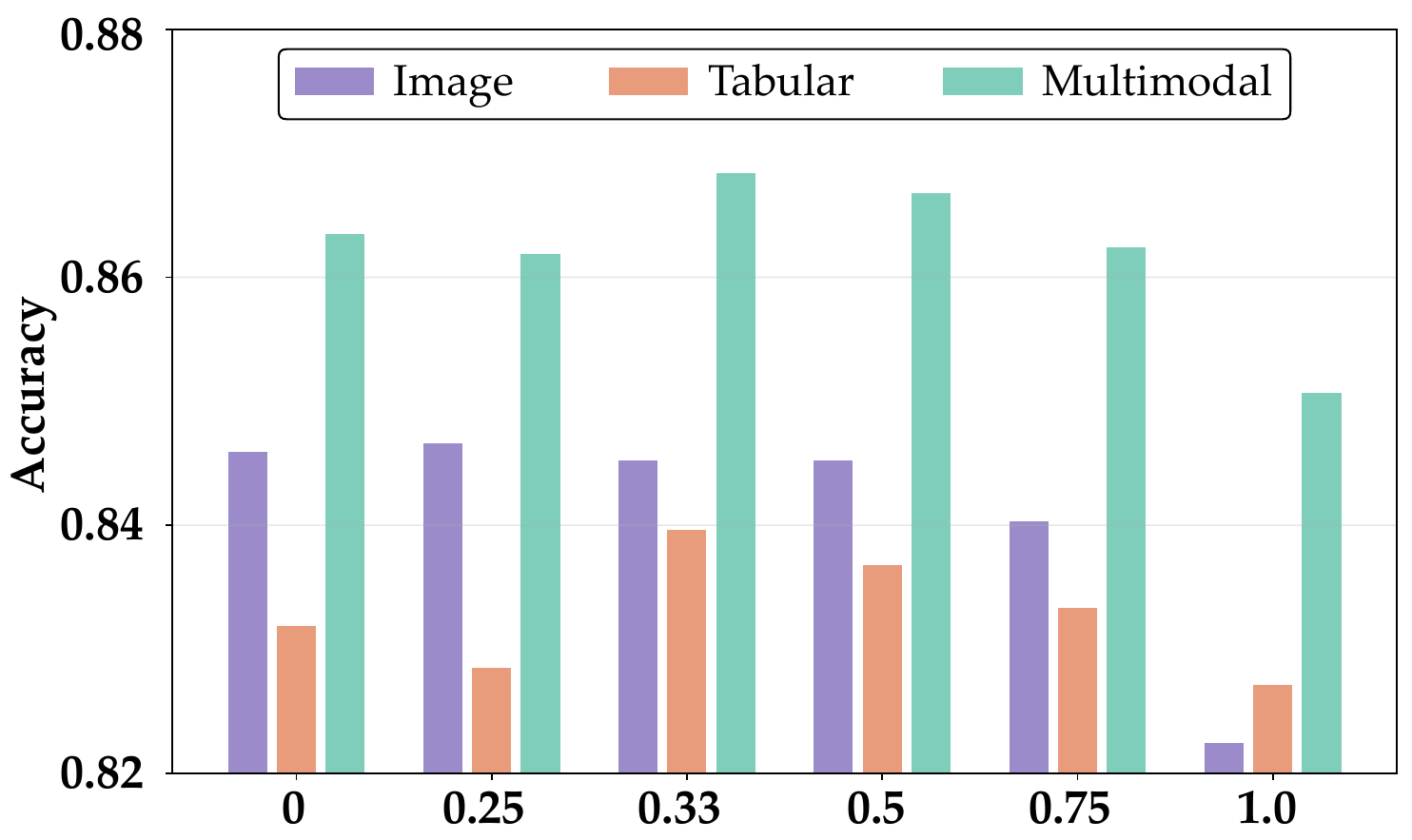}
\caption{Impact of constraint margin $\epsilon$.}
\label{fig:epsilon}
\end{figure}

\subsection{Convergence}
We also present the convergence results of the model during training on the DVM and SUNAttribute datasets. The loss curves are shown in Figure \ref{fig:convergence}. Since gradients from different modalities naturally decouple, the loss curves of unimodal consistently decrease without getting stuck.

\subsection{Impact of constraint margin $\epsilon$}
We vary $\epsilon$ on the SUNAttribute dataset and summarize the results in Figure \ref{fig:epsilon}. When $\epsilon$ is non-negative, it encourages modality gradient alignment. In particular, setting $\epsilon=0$ may induce gradient orthogonality, leading to independent updates that are suboptimal for cross-modal interaction. Our results show that properly adjusting $\epsilon$ improves performance.

\section{Conclusion}
In this work, we propose GAAL, a novel tabular-image fusion paradigm designed to address multimodal gradient conflicts and facilitate interaction. Our approach adopts alternating learning with a shared classifier to decouple multimodal gradient. To further address gradient conflicts in interaction layer and improve interaction, we design uncertainty-based cross-modal gradient surgery, which utilizes cross-modal gradients to guide the optimization of current modality. Experimental results demonstrate that GAAL outperforms all methods on both image-tabular fusion and test-time tabular missing tasks. We hope this work motivates future research on multimodal challenges encountered in real-world scenarios, with a particular focus on tabular-image learning. 

\noindent{\textbf{Limitations:}} Our method focuses on  tabular and image classification. Future work will extend the approach to regression and detection tasks.

\section*{Acknowledgment}
This work was supported in part by the NSFC (62276131), and in part by the Natural Science Foundation of Jiangsu Province of China under Grant (BK20240081).

\bibliographystyle{IEEEtran}
\bibliography{IEEEexample}

@String(IJCV = {Int. J. Comput. Vis.})

@String(CVPR= {IEEE Conf. Comput. Vis. Pattern Recog.})

@String(ICCV= {Int. Conf. Comput. Vis.})

@String(ECCV= {Eur. Conf. Comput. Vis.})

@String(ICME = {Int. Conf. Multimedia and Expo})

@String(ICLR = {Int. Conf. Learn. Represent.})

@String(IJCAI = {IJCAI})

@String(AAAI = {AAAI})

@String(IJCV  = {IJCV})

@String(CVPR  = {CVPR})

@String(ICCV  = {ICCV})

@String(ECCV  = {ECCV})

@String(ICME  =	{ICME})

@String(ICLR  = {ICLR})

@inproceedings{TIP:conf/eccv/DuZWBOQ24,
  author       = {Siyi Du and
                  Shaoming Zheng and
                  Yinsong Wang and
                  Wenjia Bai and
                  Declan P. O'Regan and
                  Chen Qin},
  title        = {{TIP:} Tabular-Image Pre-training for Multimodal Classification with
                  Incomplete Data},
  booktitle    = {ECCV},
  pages        = {478--496},
  publisher    = {Springer},
  year         = {2024},
}

@inproceedings{DI-MML:conf/mm/FanXWLG24,
  author       = {Yunfeng Fan and
                  Wenchao Xu and
                  Haozhao Wang and
                  Junhong Liu and
                  Song Guo},
  title        = {Detached and Interactive Multimodal Learning},
  publisher    = {{ACM}},
  booktitle    = {ACM MM},
  year         = {2024},
}

@article{KD:journals/corr/HintonVD15,
  author       = {Geoffrey E. Hinton and
                  Oriol Vinyals and
                  Jeffrey Dean},
  title        = {Distilling the Knowledge in a Neural Network},
  journal      = {CoRR},
  volume       = {abs/1503.02531},
  year         = {2015},
}

@inproceedings{FMR:conf/aaai/YangZFJZ17,
  author       = {Yang Yang and
                  De{-}Chuan Zhan and
                  Ying Fan and
                  Yuan Jiang and
                  Zhi{-}Hua Zhou},
  editor       = {Satinder Singh and
                  Shaul Markovitch},
  title        = {Deep Learning for Fixed Model Reuse},
  booktitle    = {Proceedings of the {AAAI} Conference on Artificial Intelligence},
  pages        = {2831--2837},
  publisher    = {{AAAI} Press},
  year         = {2017},
}

@inproceedings{MLA:conf/cvpr/ZhangYBY24,
  author       = {Xiaohui Zhang and
                  Jaehong Yoon and
                  Mohit Bansal and
                  Huaxiu Yao},
  title        = {Multimodal Representation Learning by Alternating Unimodal Adaptation},
  booktitle    = {CVPR},
  publisher    = {IEEE},
  pages        = {27456--27466}, 
  year         = {2024},
}

@inproceedings{OGR-GB:conf/cvpr/WangTF20,
  author       = {Weiyao Wang and
                  Du Tran and
                  Matt Feiszli},
  title        = {What Makes Training Multi-Modal Classification Networks Hard?},
  publisher    = {IEEE},
  booktitle    = {CVPR},
  pages        = {12692--12702},
  year         = {2020},
}

@inproceedings{OGM:conf/cvpr/PengWD0H22,
  author       = {Xiaokang Peng and
                  Yake Wei and
                  Andong Deng and
                  Dong Wang and
                  Di Hu},
  title        = {Balanced Multimodal Learning via On-the-fly Gradient Modulation},
  publisher    = {IEEE},
  booktitle    = {CVPR},
  pages        = {8228--8237},
  year         = {2022},
}

@inproceedings{ResNet:conf/cvpr/HeZRS16,
  author       = {Kaiming He and
                  Xiangyu Zhang and
                  Shaoqing Ren and
                  Jian Sun},
  title        = {Deep Residual Learning for Image Recognition},
  publisher    = {IEEE},
  booktitle    = {CVPR},
  pages        = {770--778},
  year         = {2016},
}

@inproceedings{MMCL:conf/cvpr/HagerMR23,
  author       = {Paul Hager and
                  Martin J. Menten and
                  Daniel Rueckert},
  title        = {Best of Both Worlds: Multimodal Contrastive Learning with Tabular
                  and Imaging Data},
  booktitle    = {CVPR},
  pages        = {23924--23935},
  publisher    = {IEEE},
  year         = {2023},
}

@InProceedings{Nash:conf/iccv/Wu_2025_ICCV,
    author    = {Jing Wu and Mehrtash Harandi},
    title     = {MUNBa: Machine Unlearning via Nash Bargaining},
    publisher = {{IEEE}},
    booktitle = {ICCV},
    pages     = {4754-4765},
    year      = {2025},
}

@inproceedings{ReconBoost:conf/icml/CongHua24,
  author       = {Cong Hua and Qianqian Xu and Shilong Bao and Zhiyong Yang and Qingming Huang},
  title        = {ReconBoost: Boosting Can Achieve Modality Reconcilement},
  publisher    = {PMLR},
  booktitle    = {ICML},
  year         = {2024},
}

@inproceedings{MMPareto:conf/icml/WeiH24,
  author       = {Yake Wei and Di Hu},
  title        = {MMPareto: Boosting Multimodal Learning with Innocent Unimodal Assistance},
  booktitle    = {ICML},
  publisher    = {{PMLR}},
  year         = {2024},
}

@inproceedings{CHARMS:conf/icml/JiangYW00Z24,
  author       = {Jun{-}Peng Jiang and
                  Han{-}Jia Ye and
                  Leye Wang and
                  Yang Yang and
                  Yuan Jiang and
                  De{-}Chuan Zhan},
  title        = {Tabular Insights, Visual Impacts: Transferring Expertise from Tables
                  to Images},
  booktitle    = {ICML},
  publisher    = {PMLR},
  year         = {2024},
}

@inproceedings{MFH:conf/iclr/XueGRZ23,
  author       = {Zihui Xue and
                  Zhengqi Gao and
                  Sucheng Ren and
                  Hang Zhao},
  title        = {The Modality Focusing Hypothesis: Towards Understanding Crossmodal
                  Knowledge Distillation},
  booktitle    = {ICLR},
  publisher    = {OpenReview},
  year         = {2023},
}

@inproceedings{AGEM:conf/iclr/ChaudhryRRE19,
  author       = {Arslan Chaudhry and
                  Marc'Aurelio Ranzato and
                  Marcus Rohrbach and
                  Mohamed Elhoseiny},
  title        = {Efficient Lifelong Learning with {A-GEM}},
  booktitle    = {ICLR},
  publisher    = {OpenReview},
  year         = {2019},
}

@inproceedings{KMG:conf/icmcs/PanY25,
  author       = {Hongpeng Pan and
                  Yang Yang},
  title        = {Coordinated Uni-modal Assistance for Enhancing Multi-modal Learning},
  booktitle    = {ICME},
  pages        = {1--6},
  publisher    = {{IEEE}},
  year         = {2025},
}

@inproceedings{KMG:conf/ijcai/JiangCY25,
  author       = {Qing{-}Yuan Jiang and
                  Zhouyang Chi and
                  Yang Yang},
  title        = {Interactive Multimodal Learning via Flat Gradient Modification},
  booktitle    = {IJCAI},
  pages        = {5489--5497},
  publisher    = {ijcai.org},
  year         = {2025},
}

@inproceedings{Tabular:conf/nips/GorishniyRKB21,
  author       = {Yury Gorishniy and
                  Ivan Rubachev and
                  Valentin Khrulkov and
                  Artem Babenko},
  title        = {Revisiting Deep Learning Models for Tabular Data},
  booktitle    = {NeurIPS},
  pages        = {18932--18943},
  year         = {2021},
}

@inproceedings{LFM:conf/nips/0074WJ024,
  author       = {Yang Yang and
                  Fengqiang Wan and
                  Qing{-}Yuan Jiang and
                  Yi Xu},
  title        = {Facilitating Multimodal Classification via Dynamically Learning Modality
                  Gap},
  booktitle    = {NeurIPS},
  year         = {2024},
}

@inproceedings{Tabular:conf/nips/MargeloiuJSJ24,
  author       = {Andrei Margeloiu and
                  Xiangjian Jiang and
                  Nikola Simidjievski and
                  Mateja Jamnik},
  title        = {TabEBM: {A} Tabular Data Augmentation Method with Distinct Class-Specific
                  Energy-Based Models},
   booktitle    = {NeurIPS},
  year         = {2024},
}

@inproceedings{GEM:conf/nips/Lopez-PazR17,
  author       = {David Lopez{-}Paz and
                  Marc'Aurelio Ranzato},
  title        = {Gradient Episodic Memory for Continual Learning},
  booktitle    = {NeurIPS},
  pages        = {6467--6476},
  year         = {2017},
}

@inproceedings{AUG:conf/nips/Jiang2025aug,
  author       = {Qing{-}Yuan Jiang and
                  Longfei Huang and 
                  Yang Yang},
  title        = {Rethinking Multimodal Learning from the Perspective of Mitigating Classification Ability Disproportion},
  booktitle    = {NeurIPS},
  year         = {2025},
}

@article{KMG:journals/pami/YangPJXT25,
  author       = {Yang Yang and
                  Hongpeng Pan and
                  Qing{-}Yuan Jiang and
                  Yi Xu and
                  Jinhui Tang},
  title        = {Learning to Rebalance Multi-Modal Optimization by Adaptively Masking
                  Subnetworks},
  journal      = {TPAMI},
  volume       = {47},
  number       = {6},
  pages        = {4553--4566},
  year         = {2025},
}

@inproceedings{Recommender:conf/kdd/BaltescuCPZLR22,
  author       = {Paul Baltescu and
                  Haoyu Chen and
                  Nikil Pancha and
                  Andrew Zhai and
                  Jure Leskovec and
                  Charles Rosenberg},
  title        = {ItemSage: Learning Product Embeddings for Shopping Recommendations
                  at Pinterest},
  booktitle    = {KDD},
  pages        = {2703--2711},
  publisher    = {{ACM}},
  year         = {2022},
}

@inproceedings{Healthcare:conf/kdd/ZhangCMZWWZ22,
  author       = {Chaohe Zhang and
                  Xu Chu and
                  Liantao Ma and
                  Yinghao Zhu and
                  Yasha Wang and
                  Jiangtao Wang and
                  Junfeng Zhao},
  title        = {M3Care: Learning with Missing Modalities in Multimodal Healthcare
                  Data},
  booktitle    = {KDD},
  pages        = {2418--2428},
  publisher    = {{ACM}},
  year         = {2022},
}

@article{Healthcare:acosta2022multimodal,
  title={Multimodal biomedical AI},
  author={Acosta, Juli{\'a}n N and Falcone, Guido J and Rajpurkar, Pranav and Topol, Eric J},
  journal={Nature medicine},
  volume={28},
  number={9},
  pages={1773--1784},
  year={2022},
  publisher={Nature}
}

@article{CF:spasov2019parameter,
  title={A parameter-efficient deep learning approach to predict conversion from mild cognitive impairment to Alzheimer's disease},
  author={Spasov, Simeon and Passamonti, Luca and Duggento, Andrea and Lio, Pietro and Toschi, Nicola and Alzheimer's Disease Neuroimaging Initiative and others},
  journal={Neuroimage},
  volume={189},
  pages={276--287},
  year={2019},
  publisher={Elsevier}
}

@article{DAFT:journals/neuroimage/WolfPWIa22,
  author       = {Tom Nuno Wolf and
                  Sebastian P{\"{o}}lsterl and
                  Christian Wachinger and
                  Alzheimer's Disease Neuroimaging Initiative and
                  Australian Imaging Biomarkers Lifestyle flagship study of ageing},
  title        = {{DAFT:} {A} universal module to interweave tabular data and 3D images
                  in CNNs},
  journal      = {NeuroImage},
  volume       = {260},
  pages        = {119505},
  year         = {2022},
}

@article{MF:vale2021long,
  title={Long-term cancer survival prediction using multimodal deep learning},
  author={Vale-Silva, Lu{\'\i}s A and Rohr, Karl},
  journal={Scientific Reports},
  volume={11},
  number={1},
  pages={13505},
  year={2021},
  publisher={Nature}
}

@article{Tabular:journals/tnn/BorisovLSHPK24,
  author       = {Vadim Borisov and
                  Tobias Leemann and
                  Kathrin Se{\ss}ler and
                  Johannes Haug and
                  Martin Pawelczyk and
                  Gjergji Kasneci},
  title        = {Deep Neural Networks and Tabular Data: {A} Survey},
  journal      = {TNNLS},
  volume       = {35},
  number       = {6},
  pages        = {7499--7519},
  year         = {2024},
}

@inproceedings{DVM:conf/bigdataconf/HuangCLYO22,
  author       = {Jingmin Huang and
                  Bowei Chen and
                  Lan Luo and
                  Shigang Yue and
                  Iadh Ounis},
  title        = {{DVM-CAR:} {A} Large-Scale Automotive Dataset for Visual Marketing
                  Research and Applications},
  booktitle    = {Big Data},
  pages        = {4140--4147},
  publisher    = {{IEEE}},
  year         = {2022},
}

@article{SUN:journals/ijcv/PattersonXSH14,
  author       = {Genevieve Patterson and
                  Chen Xu and
                  Hang Su and
                  James Hays},
  title        = {The {SUN} Attribute Database: Beyond Categories for Deeper Scene Understanding},
  journal      = {IJCV},
  volume       = {108},
  number       = {1-2},
  pages        = {59--81},
  year         = {2014},
}

@inproceedings{CelebA:conf/iccv/LiuLWT15,
  author       = {Ziwei Liu and
                  Ping Luo and
                  Xiaogang Wang and
                  Xiaoou Tang},
  title        = {Deep Learning Face Attributes in the Wild},
  publisher    = {{IEEE}},
  booktitle    = {ICCV},
  pages        = {3730--3738},
  year         = {2015},
}

@article{QP:frank1956algorithm,
  title={An algorithm for quadratic programming},
  author={Frank, Marguerite and Wolfe, Philip and others},
  journal={Naval research logistics quarterly},
  volume={3},
  number={1-2},
  pages={95--110},
  year={1956},
}

\end{document}